\documentclass{article}
\usepackage{spconf,amsmath,graphicx}
\usepackage{color}

\newcommand{\imageLine}[1] {
\includegraphics[trim=0pt 0pt 0pt 0pt, clip,width=0.17\textwidth]{#1colored.png}  &
\includegraphics[trim=0pt 0pt 0pt 0pt, clip,width=0.17\textwidth]{#1clipped.png}&
\includegraphics[trim=0pt 0pt 0pt 0pt, clip,width=0.17\textwidth]{#1net.png}
        \\
}

\DeclareRobustCommand\onedot{\futurelet\@let@token\@onedot}
\def\@onedot{\ifx\@let@token.\else.\null\fi\xspace}

\newcommand{\etal}{\textit{et al}. }
\title{IMAGE DECLIPPING WITH DEEP NETWORKS}
\name{Shachar Honig \& Michael Werman }
\address{Department of Computer Science, The Hebrew University of Jerusalem} 
\usepackage{titlesec}
\titlespacing*{\section}{0pt}{1.5ex plus .2ex minus .2ex}{1ex plus .2ex}
\titlespacing*{\subsection}{0pt}{1.5ex plus .2ex minus .2ex}{0.8ex plus .2ex}
\titlespacing*{\subsubsection}{0pt}{1ex plus .2ex minus .2ex}{.5ex plus .1ex}

\begin{document}
%
\maketitle
\begin{abstract}
We present a deep network to recover pixel values lost to clipping.
The clipped area of the image  is typically a uniform area of  minimum or maximum brightness, losing  image detail and color fidelity. The degree to which the clipping is visually noticeable depends on the amount by which values were clipped, and the extent of the clipped area. Clipping may occur in any (or all) of the pixel's color channels.
Although clipped pixels are common and occur to some degree in almost every image we tested, current automatic solutions have only partial success in repairing clipped pixels and work only in limited cases such as only with overexposure (not under-exposure) and when some of the color channels are not clipped. Using neural networks and their ability to model natural images allows our neural network, DeclipNet, to reconstruct data in clipped regions producing state of the art results.
\end{abstract}

\let\thefootnote\relax\footnotetext{This research was supported by the Israel Science Foundation and by the  Israel Ministry of Science and Technology.}

\section{Introduction}
\label{sec:intro}
Images often contain areas  corrupted  due to clipping.  This occurs when the scene has a higher dynamic range than the sensor. Clipping can also occur at latter stages of the image processing pipeline, for example, many cameras reduce differences in high and low intensity levels in order to expand them in the mid ranges. 
Dynamic range, the ratio between the brightest and darkest light intensities  in an outdoor scene can be 100,000:1  vs. less than 1,000:1 sensitivity for mid-range camera sensor. Even a high end camera's dynamic range is  limited compared to an outdoor scene's dynamic range. Hence, clipped pixels are common and occur to some degree in almost every image we tested. Such pixels might be clipped in one or more color channels.   
The goal of this work is to reconstruct the missing data in the over or under-saturated regions while minimizing the changes to non-saturated regions of the image. 

\begin{figure}[h] 
\hspace*{-0.72cm} 
  	\begin{tabular}{cc}
  		Clipped & Network Output \\
     	\includegraphics[trim=0 0 0 0, clip, width=1.797in]{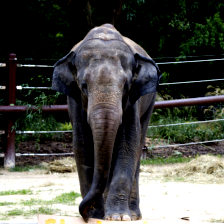} &
     	   \hspace{-4mm}
        \includegraphics[trim=0 0 0 0, clip, width=1.797in]{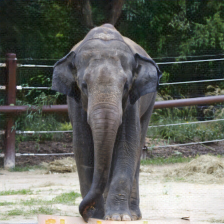} \\
     	
  	\end{tabular}
\setlength{\belowcaptionskip}{-8pt} 	 \setlength{\abovecaptionskip}{-8pt} 
\setlength{\intextsep}{-8pt} 
	\caption{Input, clipped image (left)  declipped image (right). } 
	\label{fig:exampleIntroFirst}
\end{figure}
Quantifying the difference between images has classically been measured by a pixel loss such as MSE. However using pixel loss functions in challenging problems that have many plausible solutions tend not to reconstruct high-frequency detail and produce an overly smooth result. That is  an average of the many plausible solutions. Moreover, solutions that look identical to a human might have a high Euclidean difference.
In recent work, two alternatives to pixel loss have shown excellent performance in various problems, perceptual loss and Generative Adversarial Networks (GANs). Perceptual loss has excelled in style transfer, super resolution and noise reduction  \cite{johnson_perceptual_2016}, while GANs \cite{goodfellow_generative_2014} have shown amazing results in a wide variety of problems including semantic segmentation\cite{luc_semantic_2016} and inpainting\cite{pathak_context_2016}.
\begin{figure}[t] 
\hspace*{-0.72cm} 
\setlength{\tabcolsep}{2pt}
  	\begin{tabular}{ccc}
  		Visualization& Clipped &   DeclipNet Output 
         \\ 	
  \imageLine{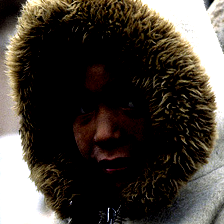}	
  \imageLine{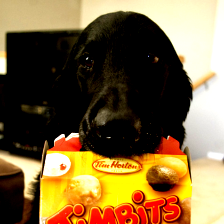}
    \end{tabular}
	\caption{\textbf{Clipped} - is input images for the DeclipNet. \textbf{Visualization} - of clipped pixels. Over-saturated pixels(max)  in all color channels were colored red and under-saturated pixels in all channels were represented in green. \textbf{DeclipNet Output} - is the result of applying the network (DeclipNet) to the clipped image} 
    
	\label{fig:exampleIntro}
\end{figure}

Classic methods for repairing saturated pixels work only where at least one color channel is non-saturated, using the non-saturated channels  to reconstruct the saturated channels. The main works were by
Zhang \etal, Masood \etal , Elhanan \etal and Guo \etal \cite{zhang_estimation_2004,masood_automatic_2009,elhanan_elboher_recovering_2010,ido_omer_color_2004,guo_correcting_2010},

These methods only try to solve over-saturation and not under-saturation and cannot reconstruct data in completely clipped regions

 Before the prevalent use of neural networks, there were other methods for image enhancement using reference images such as
\cite{joshi_personal_2010} for face images and \cite{zhang_personal_2014, savoy_recovering_2014} for famous locations.
The necessity of reference images, user assistance and  strong priors, limits the use of these techniques.

Another problem which might appear related is "single image based HDR" aka "Inverse Tone Mapping". These global methods amplify contrast, producing high-contrast high-dynamic range images but do not reconstruct missing data.
Although we did not find neural network based methods  to fix clipped images,
neural networks have surpassed previous works in a variety of challenging image transformation problems.
Image-enhancement (enhances color, does not reconstruct detail) \cite{yan_automatic_2014,gharbi_deep_2017}.
Automatic colorization (colorization of gray-scale images, does not reconstruct detail) \cite{cheng_deep_2015,zhang_colorful_2016,larsson_learning_2016}. Image inpainting, reconstructs masked areas in photos but the results are often blurry  or unrealistic, \cite{yeh_semantic_2016,radford_unsupervised_2015,pathak_context_2016}.

\begin{figure}[h] 
\includegraphics[width=0.5\textwidth]{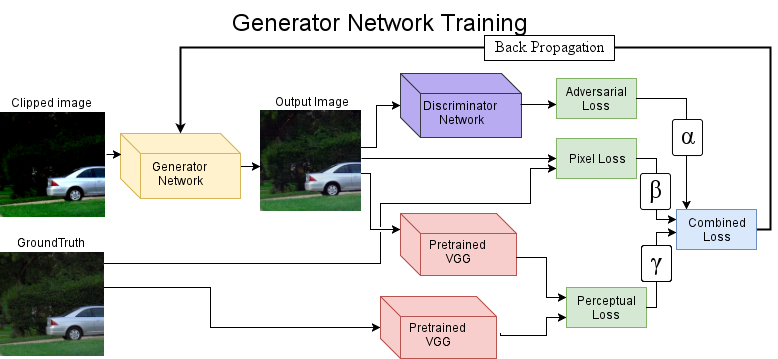}

\caption{$ \alpha$, $\beta $ and  $\gamma$  are the coefficients of the three loss functions used for the network optimization. }
\label{fig:networkgraph}
\end{figure}
\section{Contribution}
In this paper, we describe the first deep network declipper (DeclipNet). It is based on perceptual and GAN style discriminator losses for color correction and data reconstruction.
Our main contributions are:
\begin{itemize}
 \item A state of the art system to correct over and under-exposed images even when all the color channels are saturated/clipped. The network reconstructs texture and detail even in small areas with no data (all values are equal to min. or max. value)
  \item We conducted a user study that confirmed that our network's
  results look better  than previous systems, both in color correction (where not all color channels are clipped)  and in reconstructing lost data (completely saturated pixels).
\end{itemize}
We describe the network architecture in the next section. In Section \ref{Loss_Functions}, we describe the loss functions used and the technical details. A visual illustration of the results is provided in Section \ref{Results}. The paper concludes with a discussion and summary in Sections \ref{Discussion} and \ref{Summary}, respectively.

\section{Method}


We designed an image transformation neural network, DeclipNet, that takes an image with clipped pixels \(I^{c}\) and outputs an image \(I^{u}\) with the over/under-exposed pixels faithfully corrected. To train the network, we used a few image datasets and produced clipped versions of the images in them. We trained the network to restore the artificially clipped images with the original image as the ground truth.
The goal of these experiments is not to achieve state-of-the-art per pixel L2 loss between \(I^{u}\) and \(I^{o}\), but  to produce results that appear  natural and plausible as well as appearing semantically identical to \(I^{o}\) to users. The qualitative advantage of models trained using semantic reconstruction loss vs per-pixel is quite noticeable but is challenging to quantify. 

In this work we tested  the quality of the generated image using three types of loss functions:
1. pixel loss (l2), 2. perceptual loss 3. adversarial loss.  
\label{sec:method}
\begin{figure*}[h]
	\begin{tabular}{c|c}
    Generator &   Discriminator \\
    
\raisebox{-.5\height}{\includegraphics[width=0.48\textwidth]{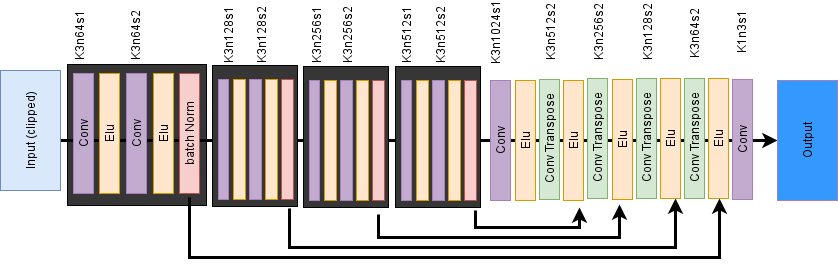}}&
\raisebox{-.5\height}{\includegraphics[width=0.48\textwidth]{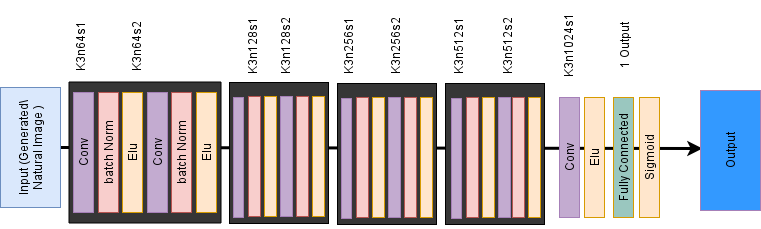}}
    \end{tabular}
\caption{The generator and discriminator architectures with corresponding kernel size (k), number of feature maps (n) and stride (s) indicated for each convolution layer.}
\label{fig:architecture}
\end{figure*}
\subsection{Network Architecture}
We explored various deep architectures   to produce an unclipped image from a clipped image. The network's output was either the entire unclipped image or only the residual between the clipped and the unclipped image.

Networks that learned the residual image were less likely to suffer from the vanishing/exploding gradient problem, learned faster and attained better results. 

In the experimentation, networks with more "skip" connections, 
inspired by ResNet\cite{he_deep_2015}, tended to produce sharper, more detailed results.

In addition, we used a third loss function, a discriminator loss. For that,  we trained a discriminator network  to distinguish between generated unclipped images and natural unclipped images. The goal of this loss function is not to compare \(I^{u}\) to \(I^{o}\) but to score how distinguishable  generated images are from  natural images. We alternated between training of the discriminator network and the generator network. For the generator adversarial loss we used the complement to the discriminator loss function. This encourages the generator to favor output images that reside on the natural image manifold.
We experimented with many different architectures, before choosing the final architecture, see Figure \ref{fig:architecture}. 
\subsubsection{Discriminator Network Architecture}
The discriminator network architecture  is  inspired by the DCGAN  class of GAN architectures proposed by Radford \etal \cite{radford_unsupervised_2015}   
We used the  exponential linear unit, Elu\cite{clevert_fast_2015} activation on all discriminator and generator layers except the last. Elu tends to lead to faster learning. The Elu activation function is the identity for positive x's and $e^x-1$ for negative x's.
We use batch-normalization in both the generator and the discriminator, and  strided convolutions in the discriminator and fractional-stride in the autoencoder.
\subsubsection{Generator Model Architecture}
The autoencoder has an hour glass shape in which all convolution and transposed convolution layers have 3x3 kernels. The activation function is Elu. In total there are 14 convolution and transposed convolution layers and the discriminator has 9 convolution layers.
\subsection{Loss Functions}\label{Loss_Functions}
\subsubsection{Mean Square Error Loss (MSE)} MSE is the standard pixel loss between  output images and the ground truth images \(|I^{u}-I^{o}|_2\) but networks trained using only MSE loss tend to be overly smooth, lacking high frequency data.
\subsubsection{Perceptual Loss} 
Building on the ideas of Gatys \etal \cite{gatys_neural_2015} and Johnson   \cite{johnson_perceptual_2016} we use a loss function that takes advantage of  networks trained on natural images  to capture the different levels of meaningful data along its processing hierarchy from pixel data at low layers to high-level
 content in the higher layers of the network. 
 We use a fully trained VGG16\cite{simonyan_very_2014} with Euclidean distance between the feature map after activation on the image \(I^{o}\) (our ground truth) and the feature map  (after activation) on the output image  \(I^{u}\) on the same layers. 
  Let \(C_{j}(I)\)  be the feature map after the j'th
 convolution layer (in our pretrained VGG16) when processing the image, I . The perceptual loss is: \({L_{2}}_{ VGG16 |(j)}(I^{o},I^{c}) =\frac{1}{ W_{j} H_{j} F_{j}} | C_{j}(I^{o})- C_{j}(I^{c}) |^2_2\)  where \( W_{j} H_{j} F_{j}\) is the size of the feature map on layer j (width, height, number of filters). Using a higher layer j  encourages the output image to be perceptually similar without forcing pixels to  match exactly.
\subsubsection{Adversarial Loss} 
The third loss function we used was an adversarial loss. This encourages the results to be on the natural image manifold and  the output tends to be crisper. 
$X=D(G(I^{c}))$, is the score of the discriminator on the output of the generator network on the input image (clipped image) with loss $max(X, 0) - X +log(1 + exp(-abs(X))$.
The discriminator learns to  distinguish between natural images and generated unclipped images. We  train the generator with  the complement to the discriminator loss.
\section{Experiments}\label{Results}

We used two popular datasets MSCOCO\cite{lin_microsoft_2014}
to train and test our networks and imageNet to validate. We trained  our networks using 300,000 images from the MSCOCO 2014 training set. For testing, we used  other MSCOCO images and imageNet images. 

We chose a random 224x224 square from every image and 
two random clipping threshold values,  overexposure between 175..255 and an underexposure threshold from 1..80. We, then, normalize the values so that they are between 0 and 1.

To train the autoencoder  we used ADAM optimization \cite{kingma_adam:_2014} with $\beta= 0.5$. 
The learning rate was
.001 with gradient clipping. For the discriminator we used SGD with a learning rate  0.1 of the generator's learning rate.
  We alternate between training the generator and training the discriminator. For the perceptual loss we found that using the last convolution layers  in  VGG16 as our feature map  worked better than using a single layer. We tried various ways to balance the three loss functions: MSE, perceptual  and adversarial  achieving best results with \\ $0.81MSE_{loss} + 0.095PER_{loss} +0.095ADV_{loss} $ 

\begin{figure}[ht] 
  	\vspace*{-0.10cm} 
 
    \setlength{\tabcolsep}{2pt}
    \begin{tabular}{ccc}
  		Clipped & Reconstructed  & Ground Truth \\
     	\includegraphics[trim=0 0 0 0, clip, width=1.097in]{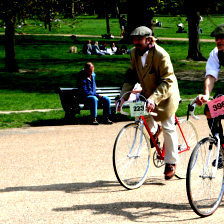} &
     	\includegraphics[trim=0 0 0 0, clip, width=1.097in]{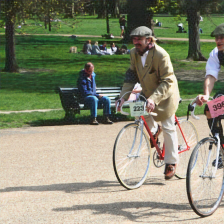}& 
        \includegraphics[trim=0 0 0 0, clip, width=1.097in]{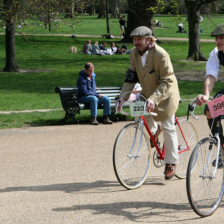} 
        \\
    \end{tabular}
	\caption{The clipped image (left),  the reconstructed image, and the unclipped ground truth image (right).}
	\label{fig:clipped_reconstructed}
\end{figure}
	 
\subsection{Evaluation}

In order to assess the perceived visual quality of the reconstruction, we performed a subjective pairwise comparison experiment between images reconstructed by our network and results of the best of the classic declipping algorithms not requiring interaction, Elboher \etal Eighteen participants, aged 21-65, took part in the experiment with approximately an equal number of men and women. 
The tasks were displayed on a 24" Dell Ips monitor.

Since Elboher's algorithm only repairs overexposure we  simulated overexposure
on 20 random images. 
The reconstructed images are reconstructions by 1. DeclipNet and 2. Elboher's  algorithm.

In each trial, the participants were shown a ground-truth image in the center and a pair of reconstructed images on either side. 
\vspace{-2mm}
\begin{figure}[h]
\hspace{-6mm}
\vspace{-2mm}
  \centering
     \fbox{\includegraphics[width=0.49\textwidth]{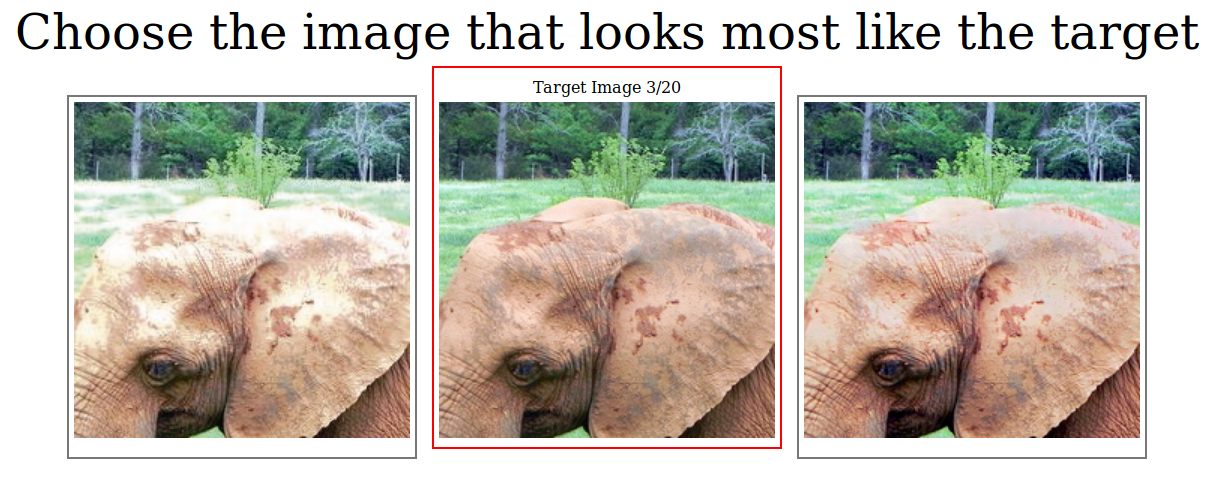}}
 \caption{ Example of a user task screen.}
\end{figure}
\vspace{-2mm}
Participants were requested to select the image that was most like the target image, had fewer artifacts and unrealistic colors,
not necessarily the prettiest image , (which might have less detail but strong contrast).
The images where presented in  random order and the side the reconstruction appeared on was random as well. 
\subsubsection{Results}
\begin{figure}[h]
  \centering
     \includegraphics[width=0.50\textwidth]{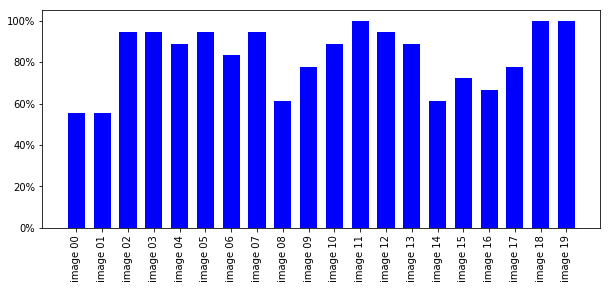}
  \caption{Preferences for the network reconstruction for the different pictures.}  
  \label{img:questions}
  \vspace{-5mm}
\end{figure}

 In all of the images the DeclipNet reconstruction was preferred,
 participants choosing it 82.5\% of the time.
The results for each picture of the image reconstruction subjective quality experiment are shown in
 Figure \ref{img:questions}.
In many cases, the images processed by DeclipNet are indistinguishable from the unclipped images.
\section{Discussion and Future Work}\label{Discussion}
The systems described in this paper substantially improve the perceptual quality of clipped images which was the goal of this work. 
The purpose of this work was to create a solution that repairs the very common over and under-exposure issues in most images and to produce results that are pleasant to the viewer and reliably create a realistic reconstruction of the original scene to a human observer. To use this solution in a modern camera or cell-phone, the network would have to be adapted to be more efficient both in memory and processing power.  Another possible adaption would be for video, which would require other considerations, e.g., consistency between frames. 

We would like to look further into the subject of developing  new types of loss functions for image content allowing better comparison between different solutions. The lack of an objective way to compare different solutions is a major challenge. Pixel loss, by itself, should not be used as a metric because an image with a substantially higher score might look inferior to a majority of viewers and vice versa.
\vspace{-2mm}
\section{Conclusion}\label{Summary}
\begin{figure}[h!] 
\vspace{-4mm}
 \centering
 \setlength{\tabcolsep}{2pt}
  	\begin{tabular}{cc}
    Clipped&DeclipNet Output\\
\includegraphics[trim=0pt 0pt 0pt 0pt, clip,width=0.23\textwidth]{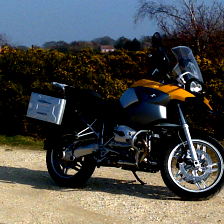}  &
\includegraphics[trim=0pt 0pt 0pt 0pt, clip,width=0.23\textwidth]{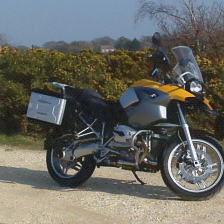}
 \end{tabular}
		\end{figure}
We presented an end-to-end trained declipping network (DeclipNet) that produces faithful and visually pleasing reconstructions of clipped images. The network augments pixel loss with perceptual loss and adversarial loss. On the basis of a user study we confirmed that DeclipNet images are significantly ”nicer” than previous solutions.


More example images can be found in the supplementary material.
\bibliographystyle{IEEEbib}
\bibliography{ref}

\end{document}